\documentclass{article}
\usepackage[preprint,nonatbib]{neurips_2022}
\usepackage[utf8]{inputenc} 
\usepackage[T1]{fontenc}   
\usepackage{hyperref}     
\usepackage{url}
\usepackage{amsfonts}
\usepackage{amssymb,amsmath,dsfont}
\usepackage{nicefrac}
\usepackage{microtype}
\usepackage{xcolor}
\usepackage{algorithm}
\usepackage{multirow}
\usepackage{graphicx}
\usepackage{bbm}
\usepackage{bm}
\usepackage{makecell}
\usepackage{wrapfig}
\usepackage{subfigure}

\newcommand{\method}[1]{\ifthenelse{\equal{#1}{full}}{Feature-Distribution Perturbation and Calibration}{PECA}}
\newcommand{\best}[1]{{\color{red}\textbf{#1}}}

\newcommand{\vparagraph}[1]{\vspace{-0em}\paragraph{#1}}

\def\ie{\emph{i.e.}, }

\def\eg{\emph{e.g.}, }

\def\etal{\emph{et al.}}
\graphicspath{{./figures/}}

\title{\method{full} \\ for Generalized ReID}

\author{
	Qilei Li \qquad Jiabo Huang \qquad Jian Hu \qquad Shaogang Gong \\ \\
	Queen Mary University of London}

\begin{document}

\maketitle
\begin{abstract}
  Person Re-identification (ReID)
  has been advanced remarkably
  over the last 10 years
  along with the rapid development of
  deep learning for visual recognition.
  However, the i.i.d. (independent and identically distributed) assumption
  commonly held in most deep learning models
  is somewhat non-applicable to ReID
  considering its objective to
  identify images of the same pedestrian
  across cameras at different locations
  often of variable and independent domain characteristics that
  are also subject to view-biased data distribution.
  In this work,
  we propose a \method{full} (\method{abbr}) method
  to derive generic feature representations for person ReID,
  which is not only discriminative across cameras
  but also agnostic and deployable to arbitrary unseen target domains.
  Specifically,
  we perform per-domain feature-distribution perturbation
  to refrain the model from overfitting to
  the domain-biased distribution of each source (seen) domain
  by enforcing feature invariance
  to distribution shifts
  caused by perturbation.
  Furthermore,
  we design a global calibration mechanism
  to align feature distributions across all the source domains
  to improve the model's generalization capacity
  by eliminating domain bias.
  These local perturbation and global calibration
  are conducted simultaneously,
  which share the same principle to avoid models overfitting
  by regularization respectively
  on the perturbed and the original distributions.
  Extensive experiments
  were conducted on eight person ReID datasets
  and the proposed \method{abbr} model
  outperformed the state-of-the-art competitors
  by significant margins.
\end{abstract}

\vspace{-1em}
\section{Introduction}
\vspace{-0.5em}
\label{sec:intro}
Person Re-identification (ReID)
aims to identify the images of the same pedestrians
captured by non-overlapping cameras at different times and locations.
It has achieved remarkable success
when both training and testing are performed in the same domains
~\cite{li2018harmonious,zheng2019joint,zhang2020relation}.
However,
the widely held i.i.d.
assumption
does not always hold in real-world ReID scenarios due to
significantly diverse viewing conditions at different locations of
biased distributions at different camera views, and more generally
across different application domains.
As a result,
a well-trained model will degrade significantly
when applied to unseen new target domains~\cite{luo2020generalizing,choi2021meta,wei2018person}.
To that end,
Domain Generalization (DG)~\cite{zhou2021domain,zhou2020learning,mahajan2021domain},
which aims at learning a domain-agnostic model,
has drawn increasing attention in the ReID community.
It is a more practical and challenging problem,
which requires no prior knowledge about the target test domain
to achieve ``out-of-the-box'' deployment.

\begin{figure}[tbp]
  \centering
  \includegraphics[width=\textwidth]{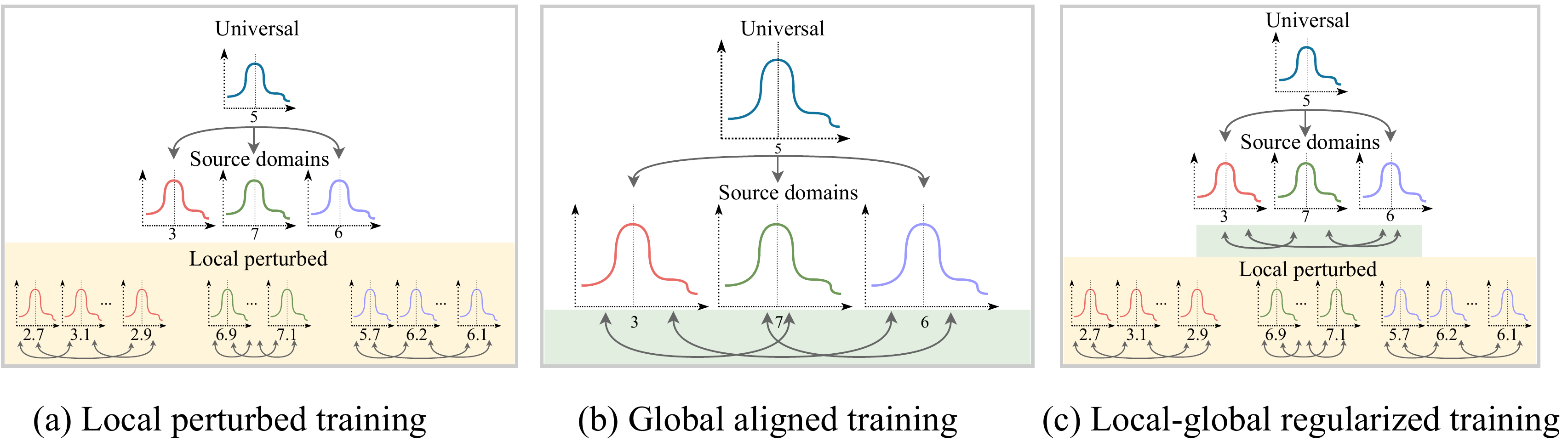}
  \vspace{-1em}
  \caption{Illustration of three training schemes
    in domain generalized ReID.
    The `Universal' is ideally the distribution
    for any new target domains.
    The source domains are differentiated by different colors,
    and the perturbed distributions share the same color
    with the corresponding original.
    The number indicates the characteristics of that distribution,
    and similar value means a smaller domain gap,
    vice versa.
    The proposed \protect\method{abbr} model
    simultaneously conducts local perturbation and global calibration
    to eliminate domain bias
    for learning a domain-agonistic representation.
  }
  \vspace{-2em}
  \label{fig:method_compare}
\end{figure}

Recent attempts on generalized ReID
aim to prevent models from overfitting
to the training data
in source domains
from either a local perspective
by manipulating the data distribution of each domain,
or in a global view
to represent the samples of all domains
in a common representational space.
The local-based methods
\cite{dai2021generalizable,yu2021multiple,jin2020style,jia2019frustratingly}
are usually implemented by
feature perturbation and/or normalization,
as shown in Figure~\ref{fig:method_compare} (a).
However,
the perturbed distributions
constructed from the original data of a single source domain
is subject to subtle distribution shift and also domain biased.
On the other hand,
the global-based approaches
~\cite{choi2021meta, ang2021dex,zhou2020learning,zhang2021learning}
aim to align the feature distributions
of multiple domains
so that the per-domain data characteristic
(\ie mean and variance of the data distribution
which is assumed to be a Gaussian distribution)
is ignored when representing images of different domains,
as illustrated in Figure~\ref{fig:method_compare} (b).
They often explicitly pre-define a target distribution to be aligned towards,
or implicitly learn a global consensus by
training a single model with data of all the source domains.
However,
even the domain gap
is reduced by such a global regularization
from restricted `true' distributions,
the learned representations are inherently domain-biased toward
the consensus of the multiple seen training domains
rather than the desired universal distribution scalable to unseen target domains
given the number of domains available for training
is always limited.

In this work,
we present a \textit{\method{full}} (\method{abbr}) model
to accomplish generalized ReID
with the objective
to learn more generalizable discriminative representations
for model deployment to unseen target domains.
This is achieved by regularizing model training
simultaneously with local distribution perturbation
and global distribution calibration,
as depicted in Figure~\ref{fig:method_compare}(c).
Specifically,
on the one hand,
as each source domain usually depicts
limited numbers of pedestrians
under certain scenarios,
simply training from such data
will lead to overfitting to
the domain-specific inherently domain biased distribution,
which harms the model's generalizability.
To address this issue,
we introduce local perturbation module
to diversify the feature distribution
based on a perturbing factor
estimated per domain,
which enables the model to be more invariant
to distribution shifts.
On the other hand,
despite the unpredictable distribution gaps
between different ReID data domains
due to the undesirable scenario-sensitive information
embedded in images specific to each domain, \eg the background,
we consider that
the features derived from different independent domains
should share a high proportion of
information as the universally applicable
explanatory factors for domain-independent identity discrimination.
In this regard,
we propose to simultaneously calibrate
the feature distributions
across all the source domains,
so to eliminate the domain-specific data characteristics
in feature representations
that are potentially caused by the identity-irrelevant redundancy.
Both the proposed local perturbation and global calibration modules
reinforce the same purpose of regularizing the model training,
but they are devised in different hierarchies
and complementary to each other.
Different from the existing methods
which consider only partially
from the local or global perspectives,
our method handles both
to promote the model in learning domain-agnostic representations.

Contributions of this work are three-fold:
\textbf{(1)} To our best knowledge,
we make the first attempt to exploit jointly
the local feature-distribution perturbation and
the global feature-distribution calibration
for improving the model's generalizability
to arbitrary unseen domains
while maintaining its discrimination.
\textbf{(2)} We formulate a local perturbation module (LPM)
to diversify per-domain feature distribution
so to refrain the model from overfitting
to each source domain,
and a global calibration module (GCM)
to further eliminate domain bias
by aligning the distribution
of multiple source domains. We simultaneously regularize both to
strike the optimal balance between these two competing objectives.
\textbf{(3)}. Extensive experiments verify
the superior generalizability
of the proposed \method{abbr} model
over the state-of-the-art DG models
on a wide range of ReID datasets
by a notable margin,
\eg the mAP is improved absolutely by 5.8\% on Market1501
and the Rank-1 is by 6.0\% on MSMT17.

\vspace{-0.5em}
\section{Related Works}
\label{sec:related}
\vspace{-0.5em}
\noindent {\bf Generalized Person ReID.}
Person ReID aims to match the same identity
across disjoint cameras.
However,
a well-trained ReID model always significantly degrades
when evaluated on novel unseen domains,
caused by the domain bias
between training and testing data.
To obtain a robust model
in achieving ``out-of-the-box'' deployment,
recent attempts on generalized ReID
avoids models from overfitting to the source domains
by either local domain manipulation
or global cross-domain alignment
to extract domain-invariant features more resistant to
domain bias.
\vspace{-1em}
\vparagraph{\em Local domain data manipulation.}
It is easy to train separate local models
with labeled samples
respective to each source domain,
with subsequently a model aggregation~\cite{dai2021generalizable, yu2021multiple}.
However,
these local models
would overfit to the corresponding source domain,
while losing generalizability to others.
A natural way to solve this problem is
either to diverse the local training samples
for learning a knowledgeable model,
or to eliminate biased information within each source domain
for learning a domain-unbiased model.
Both solutions fall into the category of local domain manipulation,
which alters the data distribution in a per-domain manner.
For data diversification,
the most intuitive approach
is to perform augmentation,
on either the raw image~\cite{chen2020self}
or feature spaces~\cite{fu2019self}.
For eliminating local domain bias,
normalization,
which regulates the data distribution
based on the data statistics,
has been widely studied recently
Jin~\etal~\cite{jin2020style} introduced
instance normalization (IN)
for restituting the style component
out of an ID representation.
Jia~\etal~\cite{jia2019frustratingly}
combined batch normalization with instance normalization
in a unified architecture
to achieve content and style unification.
However,
these local diagrams
consider only per-domain information
during feature perturbation,
and is still subject to subtle distribution shifts.
In this work,
we propose to perturb the per-domain feature distribution
to empower the model to be agnostic to holistic domain shift.
The complementary regularization
provided by the global distribution calibration remedy
helps the learned model
being invariant
against both perturbed distribution shift
and real domain gap,
so to extract generic
yet discriminative representation
for any unseen domain.
\vspace{-1em}
\vparagraph{\em Global distribution calibration.}
In contrast to the local approaches,
methods based on global distribution calibration
consider the cross-domain association
by learning a shared representational space for all domains.
These methods are built based on a straightforward assumption
that source invariant features
are also invariant to any unseen target domains~\cite{li2018domain}.
In this spirit,
DEX~\cite{ang2021dex} dynamically performed e space expansion
towards the direction of a zero-mean normal distribution
with a covariance matrix estimated from the corresponding domain.
Recent works~\cite{zhao2021learning,choi2021meta} took the idea of meta-learning
with the aim of ``learning to generalize''
by randomly splitting available source domains into meta-training and meta-testing sets,
to mimic real-world deployment scenarios.
Such a scheme implicitly aligns the cross-domain feature distributions
to a shared space by randomly setting the alignment target,
\ie the meta-testing set.
Zhang \cite{zhang2021learning} proposed learning causal invariant feature
by disentangling ID-specific and domain-specific factors
for all the training samples
from all the source domains,
which enables the disentangled feature to well-preserved ID information
while sharing the same feature space for all the domains.
However,
even aligning among multiple `real' source domains
can reduce the domain gap,
the learned representations are still biased
towards the consensus of the limited seen training domains,
instead of the desired universal distribution scalable to unseen
target domains.
In this work,
we propose to associate global alignment
with local perturbation to achieve hierarchical regularization
to avoid the model from overfitting to the source domains,
so to learn domain-agnostic representations.

\noindent {\bf Data Augmentation.}
The conventional paradigm of data augmentation
is to diversify {\em data}.
GANs~\cite{goodfellow2014generative}
have also been extensively explored to generate new {\em data} samples.
Yang~\etal~\cite{yang2021adversarial} designed an image augmentation module
which helps the network to learn domain-invariant representation
by distilling information
learned from the augmented samples
to the teacher network.
More recently, feature augmentation has emerged for semantic
transformations.
DeepAugment~\cite{hendrycks2021many} perturbed features
via stochastic operations
by forwarding images through a pre-trained image-to-image model,
to generate semantically meaningful and diverse samples.
Li~\etal~\cite{li2021simple} discovered that
embedding white Gaussian noise
in high-dimensional feature space
provides substantive statistics reflective of cross-domain variability.
Li~\etal~\cite{li2022uncertainty} proposed to model the feature uncertainty
with a multivariate Gaussian distribution
to perturb hierarchical features
to diversify the feature space.
In this work,
we explore feature distribution augmentation in each source domain
to achieve per-domain feature distribution diversification rather than
diversifying the data,
with the objective of
making the model invariant to per-domain holistic shift in order to
avoid model overfitting in each source domain.

\noindent {\bf Distribution Alignment.}
The idea of distribution alignment
aims to minimize the feature discrepancy
between source and target domains.
However,
it is impossible for DG
to explicitly perform such ``target-oriented'' alignment
due to the absence of target domains during model training.
With a straightforward assumption
that features which are invariant to the source domain shift
should also be invariant to any unseen target domain~\cite{li2018domain},
DG approaches share the spirit
to minimize the discrepancy among source domains
to achieve distribution alignment.
There are a wide variety of statistical metrics
available for minimizing,
such as Euclidean distance and $f$-divergences.
In this regard,
Li~\etal~\cite{li2020domain} proposed to minimize the KL divergence
of source domain features
with a Gaussian distribution.
Several researchers achieved distribution alignment by
minimizing a single moment (mean or variance)~\cite{muandet2013domain,hu2020domain}
or multiple moments ~\cite{erfani2016robust,ghifary2016scatter}
calculated over a batch of source domain samples
through either a projection matrix~\cite{ghifary2016scatter}
or a non-linear deep network~\cite{jin2020feature}.
Li~\etal~\cite{li2018domain} minimized the MMD distance
by aligning the source domain feature distributions
with a prior distribution
via adversarial training~\cite{goodfellow2014generative}.
In this paper,
the proposed global distribution calibration
operates on the same principle
to align the source domains
in learning a domain-agnostic model.
Differently,
we tailor the alignment objective for person ReID
considering that all samples are depicting pedestrians,
rather than predefine a  deterministic distribution to align,
\eg Gaussian or Laplace distributions.
Specifically,
we constructed a common feature space
upon the ID prototypical representations
stored in a global memory bank,
so as to eliminate domain-biased information.

\section{Balancing Feature-Distribution Local Perturbation with Global Calibration}
\label{sec:method}
Given $K$ source domains $\mathcal{D} = \{D^k\}_{k=1}^{K}$,
the objective of generalized ReID is to
derive a domain-agnostic model $\theta$
which is capable of extracting domain-invariant representations
for identity retrieval by a distance metric,
\eg Cosine similarity or Euclidean distance,
for any \textit{unseen} target domain $D^t$,
This is inherently challenging
due to the unpredictable domain gap
between training and testing data.

\begin{figure}
  \centering
  \includegraphics[width=\textwidth]{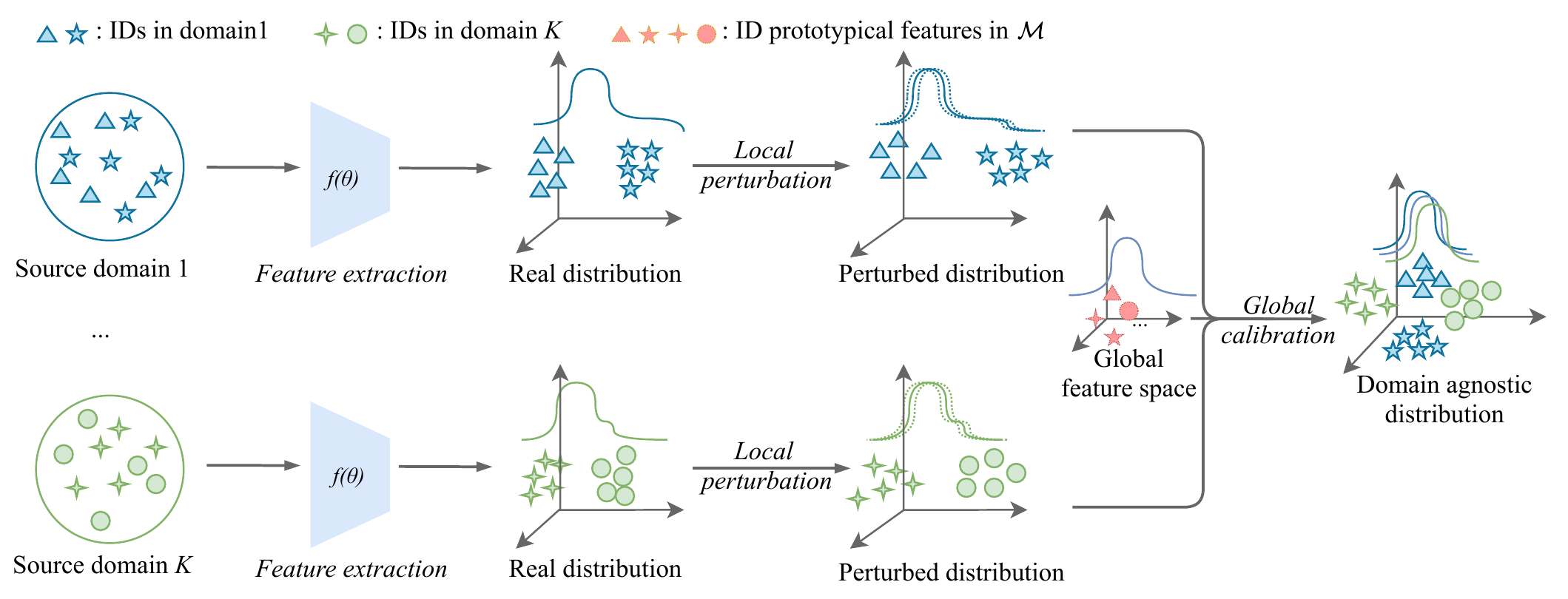}
  \caption{Overview of the proposed
    \textit{\protect\method{full}} (\protect\method{abbr}) model.
    The overall objective is to derive generic feature representation
    by avoiding model overfitting to the source domains,
    which is achieved by \textit{Local Perturbation Module}
    to enforce the learned feature invariant
    to per-domain distribution shifts caused by perturbation,
    and \textit{Global Calibration Module}
    to align cross-domain distribution
    regardless of domain annotations.
  }
  \vspace{-1em}
  \label{fig:framework}
\end{figure}
\subsection{Overview}

In this work,
we propose a \textit{\method{full}} (\method{abbr}) model
to derive domain-agnostic yet discriminative ID representations.
It regularizes the model training
to satisfy simultaneously both local perturbation
and global calibration.
The local regularization is built to perform
per-domain {\em feature-distribution} diversification,
and the global calibration is designed to achieve
cross-domain {\em feature-distribution} alignment,
as shown in Figure~\ref{fig:framework}.
During training,
for each source domain $D^k$,
a batch of samples
$({x^k}, {y^k})$
is fed into the network backbone
to extract the feature map
${e}^k$.
Then we perform per-domain diversification
with Local Perturbation Module (LPM) as
\begin{equation}
  \{{\hat{e}}^k\}_{k=1}^K = \{l({e}^k)\}_{k=1}^K,
  \label{equ:local_overview}
\end{equation}
where $l(\cdot)$ is the function of LPM
to enable the local model to be invariant
against per-domain shifts by training
with the perturbed features $\{{\hat{e}}^k\}_{k=1}^K$.

The balancing Global Calibration Module (GCM)
further regularizes the model learning
by aligning the holistic representation
(the input feature of the classifier)
into a common feature space
constructed from $\mathcal{M}$
regardless of domain label.
To distinguish the holistic representation
from the intermediate representation $e^k$,
we note it as $v^k \in \mathbb{R}^{B \times d}$
and its perturbed counterpart as $\hat{v}^k$ correspondingly,
where $d$ is a hyperparameter
to the representation dimension.
This global regularization is mathematically formulated as
\begin{equation}
  \mathcal{L}_\text{g}({\hat{v}}^k, \mathcal{M}) =
  ||\text{dist}(\hat{v}^k),\text{dist}(\mathcal{M})||_1,
  \label{equ:global_overview}
\end{equation}
where $\mathcal{L}_g(\cdot)$ is the global regularization term
aiming to align the distribution
of holistic ID representations $\text{dist}(\hat{v}^k)$
with the global distribution $\text{dist}(\mathcal{M})$.

As complementary to the LPM,
GCM focus on cross-domain regularization
by pulling representations into a domain-agnostic space,
thus empowering the generalizability of the ReID model
for any unseen novel domain.
With the collaboration of LPM and GCM,
the \method{abbr} model can be trained
with arbitrarily conventional ReID objectives
in an end-to-end manner.
When deployed to an unseen novel domain,
a generic distance metric (\eg Euclidean or Cosine distance)
is used to measure the pairwise representational similarity
between the query image against
the galleries for identity retrieval.

\subsection{Local Feature-Distribution Perturbation}
\label{sec:local}
Given an intermediate feature representation
${e}^k_{i} \in \mathbb{R}^{B \times C \times H \times W}$
extracted from the source domain ${D^k}$ at $i$-th layer,
the objective of LPM is to
perturb per-domain features
to avoid local-domain overfitting.
For notation clarity,
we omit the layer index $i$
in the following formulations.
Inspired by feature augmentation~\cite{li2021simple}
and Instance Normalization~(IN)~\cite{huang2017arbitrary, li2022uncertainty},
LPM performs perturbation
by randomly substituting the transformation factors
of IN.
Specifically,
we first calculate the channel-wise moments
$\mu ({e}^k)\in \mathbb{R}^{B \times C}$ and $\sigma ({e}^k)\in \mathbb{R}^{B \times C}$ for IN as
\begin{equation}
  \begin{gathered}
    \mu({e}^k) = \frac{1}{HW}\sum_{h=1}^{H}\sum_{w=1}^{W} {e}^k_{h,w}, \quad
    \sigma({e}^k) = \frac{1}{HW}\sum_{h=1}^{H}\sum_{w=1}^{W}({e}^k_{h,w}-\mu({e}^k))^2. \
  \end{gathered}
\end{equation}
As suggested by~\cite{jin2020style},
these statistical moments encode
not only style information
but also certain task-relevant information
dedicated to ReID.
Instead of discarding all of them
for style bias reduction
as adopted in~\cite{jia2019frustratingly, zhao2021learning},
we propose to maintain the discrimination
while increasing the local-domain data diversity
by holistically shifting its distribution.
This is achieved by perturbing the per-domain instance moments as
\begin{equation}
  \begin{gathered}
    \hat{\mu}({e}^k) = \mu({e}^k) + \epsilon_{\mu} \text{h}(\mu({e}^k)), \quad
    \hat{\sigma}({e}^k) = \sigma({e}^k) + \epsilon_{\sigma} \text{h}(\sigma({e}^k)),
  \end{gathered}
\end{equation}
where $\text{h}(\cdot)$ calculate the perturbation factors,
which are mathematically the standard derivation.
They reflect the dispersed level of the local domain,
and ensures the perturbation within a plausible range,
so to avoid over-perturbation
which causes model collapse,
or under-perturbation
which cannot provide any benefit in model learning.
$\epsilon_\mu$ and $\epsilon_\sigma$
varies the perturbation intensity
to guarantee the diversity of perturbed features,
and both are randomly sampled from a standard normal distribution.
We subsequently perform feature transformation
by substituting the local-domain moments as
\begin{equation}
  \begin{gathered}
    {\hat{e}}^k = \hat{\sigma}({e}^k) \frac{{e}^k-\mu({e}^k)}
    {\sigma({e}^k)} + \hat{\mu}({e}^k).
  \end{gathered}
  \label{equ:lpm}
\end{equation}
By introducing the perturbed representation ${\hat{e}}^k$,
the per-domain feature becomes more diverse
so to improve the model's generalizability
against the per-domain shift.

\subsection{Global Feature-Distribution Calibration}
\label{scr:glbal}
The global calibration module (GCM) is complementary to LPM
by aligning the distribution of cross-domain features
into a common feature space.
GCM considers the association
between the perturbed holistic representation $\hat{v}^k$
and a global memory bank $\mathcal{M}$.
Specifically,
we calculate the global statistical moments
${\mu_\text{g}} \in \mathbb{R}^d$
and ${\sigma_\text{g}} \in \mathbb{R}^d$
in each training iteration as
\begin{equation}
  \begin{gathered}
    {\mu_\text{g}} = \frac{1}{K}\frac{1}{N^k}\sum_{k=1}^{K}\sum_{n=1}^{N^k}{\mathcal{M}_n^k}, \quad
    {\sigma_\text{g}} = \frac{1}{K}\sum_{k=1}^{K}\sum_{n=1}^{N^k}(\mathcal{M}{_n^k} - {\mu_g}),
  \end{gathered}
\end{equation}
where $\mathcal{M}^k_n \in \mathbb{R}^{d}$
is the prototypical feature of the $n$-th identity
in the $k$-th domain.
These global statistical moments depict a feature space
shared by the prototypical representations on $\mathcal{M}$ for all the identities.
Subsequently,
the holistic representations are calibrated
into the joint feature space by
\begin{equation}
  \mathcal{L}_\text{g}({\hat{v}}^k, \mathcal{M}) =
  \frac{1}{K} \sum_{k=1}^K (||\mu({\hat{v}}^k) - {\mu_\text{g}}||_1 +
  ||\sigma({\hat{v}}^k) - {\sigma_\text{g}}||_1).
  \label{equ:gcm}
\end{equation}
Here, $\mu({\hat{v}}^k) \in \mathbb{R}^d$
and $\sigma({\hat{v}}^k) \in \mathbb{R}^d$
are the channel-wise mean and standard derivation
of the perturbed representation ${\hat{v}}^k$.
GCM enables the extracted features
to fall into a domain-invariant space.
The hierarchical regularization
achieved by LPM and GCM
makes the model generic
in extracting domain-agnostic representations.

\subsection{Training Pipeline}
\vparagraph{Learning objective.}
Given the formulations of LPM and GCM,
the proposed \method{abbr} can benefit from
conventional learning supervisions.
Specifically,
the \method{abbr} model is jointly trained with
a softmax cross-entropy loss $\mathcal{L}_\text{id}$
and the global regularization item $\mathcal{L}_\text{g}$ as
\begin{equation}
  \begin{gathered}
    \mathcal{L} = \mathcal{L}_\text{id} + \lambda \mathcal{L}_\text{g},\quad
    \mathcal{L}_\text{id}(x^k, y^k) =
    - \sum_{j=1}^C p^k_{j} \log \tilde{p}^k_{j},\quad
    \tilde{p}^k = \text{Softmax}(\text{MC}({\hat{v}}^k)).
  \end{gathered}
  \label{equ:loss_all}
\end{equation}
The notations $x^k$ and $y^k$ are
the raw input images sampled from domain $D^k$
and its corresponding ID label,
respectively,
whilst
$p^k$ is a one-hot distribution
activated at $y^k$.
The function $\text{MC}(\cdot)$ stands for the
memory-based classifier~\cite{zhong2019invariance, zhao2021learning},
and $\lambda$ decides
the importance of $\mathcal{L}_\text{g}$
regarding the identity loss $\mathcal{L}_\text{id}$.

\vparagraph{Memory bank update.}
In each training iteration,
once the network parameters are updated
according to $\mathcal{L}$ (Eq.~\eqref{equ:loss_all}),
the memory bank $\mathcal{M}$
is then refreshed
by Exponential Moving Average (EMA) as
\begin{equation}
  M^k_{y^k} = \beta M^k_{y^k} + (1 - \beta) \hat{v}^k, \quad k = \{1, \dots, K\},
  \label{equ:mem_ema}
\end{equation}
in which $\beta$ is the EMA momentum.
The prototypical features
in the memory bank $\mathcal{M}$ is iteratively updated
with the latest corresponding ID representations.
Consequently,
a more discriminative feature space
will be yielded by $\mathcal{M}$ for global alignment.

\section{Experiments}
\label{sec:exp}

\begin{figure}[ht]
  \centering
  \includegraphics[width=\textwidth]{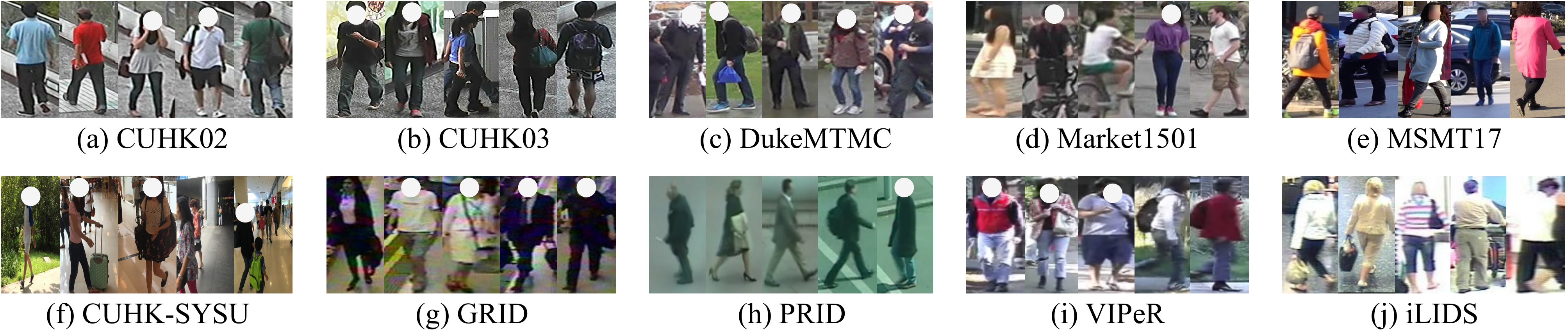}
  \vspace{-1.5em}
  \caption{Example identity samples from different domains.
    Significant domain gaps are caused by the variation
    on nationality, illumination, viewpoints, resolution, scenario, etc.}
  \vspace{-1.5em}
  \label{fig:dataset_example}
\end{figure}

\vparagraph{Datasets and protocols.}
We conducted multi\-source domain generalization
on a wide-range of benchmarks,
including Market1501 (M)~\cite{zheng2015scalable},
DukeMTMC (D)~\cite{zheng2017unlabeled},
MSMT17 (MT)~\cite{wei2018person},
CUHK02 (C2)~\cite{li2013locally},
CUHK03 (C3)~\cite{li2014deepreid},
CUHK-SYSU (CS)~\cite{xiao2016end},
and four small datasets including
PRID~\cite{hirzer2011person},
GRID~\cite{loy2010time},
VIPer~\cite{gray2008viewpoint},
and iLIDs~\cite{zheng2009associating}.
The statistics of these datasets are shown in Table~\ref{tab:data_stat},
and a few samples are visualized in Figure~\ref{fig:dataset_example},
which reveals significant domain gaps caused by variations
on nationality, illumination, viewpoint, resolution, scenario, etc.
Mean average precision (mAP)
and CMC accuracy
on Rank-1
are adopted as evaluation metrics.

\begin{wraptable}{r}{0.5\linewidth}
  \vspace{-0.6em}
  \centering
  \caption{Statistics of ReID datasets.
  }
  \resizebox{0.5\textwidth}{!}{
    \begin{tabular}{lcccc}
    \hline
     & \multicolumn{2}{c}{Probe} & \multicolumn{2}{c}{Gallery}\\
    \cline{2-5}
    \multirow{-2}{*}{Datasets} & ID & Img  & ID & Img\\
    \hline
    PRID~\cite{hirzer2011person} & 100 & 100 & 649 & 649\\
    GRID~\cite{loy2010time} & 125 & 125 & 900 & 900\\
    VIPeR~\cite{gray2008viewpoint} & 316 & 316 & 316 & 316\\
    iLIDS~\cite{zheng2009associating} & 60 & 60 & 60 & 60\\
    \hline\hline
     & \multicolumn{2}{c}{Abbr.} & ID & Img\\
    \hline
    Market1501~\cite{zheng2015scalable} & \multicolumn{2}{c}{M} & 1,501 & 29,419\\
    DukeMTMC~\cite{zheng2017unlabeled} & \multicolumn{2}{c}{D} & 1,812 & 36,411\\
    MSMT17~\cite{wei2018person} & \multicolumn{2}{c}{MS} & 4,101 & 126,441\\
    CUHK02~\cite{li2013locally} & \multicolumn{2}{c}{C2} & 1,816 & 7,264\\
    CUHK03~\cite{li2014deepreid} & \multicolumn{2}{c}{C3} & 1,467 & 14,097\\
    CUHK-SYSU~\cite{xiao2016end} & \multicolumn{2}{c}{CS} & 11,934 & 34,574\\
    \hline
    \end{tabular}
  }
  \label{tab:data_stat}
  \vspace{-1em}
\end{wraptable}

\vspace{-0.5em}
\vparagraph{Implementation details.}
We used ResNet50~\cite{he2016deep}
pre-trained on ImageNet to bootstrap our feature extractor.
The batch size was set to 128,
including 16 identities
and 8 images for each.
All images were resized to $256 \times 128$.
We randomly augmented the training data
by cropping, flipping, and colorjitter.
The proposed \method{abbr} was trained 60 epochs
by Adam optimizor~\cite{kingma2014adam},
and we adopted the warm-up strategy in the first 10 epochs
to stabilize model training.
The learning rate was initialized as $3.5e-4$
and multiplied by 0.1 at 30th and 50th epoch.
The momentum for the memory update was set to 0.8.
The dimension of extracted representations
was conventionally set to 2048.
All the experiments were conducted
on the PyTorch~\cite{paszke2017automatic} framework
with four A100 GPUs.

\vspace{-0.5em}
\subsection{Comparisons to the State-Of-The-Art}
\label{sec:sota}
\vspace{-0.5em}
\vparagraph{Comparison under the traditional benchmark setting.}
Under the existing benchmark setting
\cite{dai2021generalizable, jin2020style,song2019generalizable},
five datasets (M+D+C2+C3+CS, as in Table~\ref{tab:data_stat}) were used as source domains,
and the generalizability was evaluated on four {\em small-scale}
datasets of different domains not contributing to training (unseen),
which are PRID, GRID, VIPeR, iLIDs.
All the images in the source domains
were used for training,
without the original training or testing splits.
Being consistent with existing performance evaluation protocols~\cite{jin2020style,song2019generalizable},
we performed 10-trail evaluations
by randomly splitting query/gallery sets,
and reported the averaged performance in Table~\ref{tab:sota_small},
which shows the considerable superiority of the proposed \method{abbr}
over the state-of-the-art (SOTA) competitors.

\begin{table}[ht]
  \centering
  \caption{Comparisons with the SOTA methods under traditional setting.
    Best results are in \best{bold}.
  }
  \resizebox{\textwidth}{!}{
    \begin{tabular}{l|c|c|c|c|c|c|c|c|c|c}
\hline
 & \multicolumn{2}{c|}{PRID} & \multicolumn{2}{c|}{GRID} & \multicolumn{2}{c|}{VIPeR} & \multicolumn{2}{c|}{iLIDs} & \multicolumn{2}{c}{Average}\\ 
\cline{2-11}
\multirow{-2}{*}{Method} & mAP & Rank-1 & mAP & Rank-1 & mAP & Rank-1 & mAP & Rank-1 & mAP & Rank-1\\ 
\hline
Agg\-Align~\cite{zhang2017alignedreid} & 25.5 & 17.2 & 24.7 & 15.9 & 52.9 & 42.8 & 74.7 & 63.8 & 44.5 & 34.9\\ 
Reptile~\cite{nichol2018first} & 26.9 & 17.9 & 23.0 & 16.2 & 31.3 & 22.1 & 67.1 & 56.0 & 37.1 & 28.0\\ 
CrossGrad~\cite{shankar2018generalizing} & 28.2 & 18.8 & 16.0 & 9.0 & 30.4 & 20.9 & 61.3 & 49.7 & 34.0 & 24.6\\ 
Agg\_PCB~\cite{sun2019learning} & 32.0 & 21.5 & 44.7 & 36.0 & 45.4 & 38.1 & 73.9 & 66.7 & 49.0 & 40.6\\ 
MLDG~\cite{li2018learning} & 35.4 & 24.0 & 23.6 & 15.8 & 33.5 & 23.5 & 65.2 & 53.8 & 39.4 & 29.3\\ 
PPA~\cite{qiao2018few} & 45.3 & 31.9 & 38.0 & 26.9 & 54.5 & 45.1 & 72.7 & 64.5 & 52.6 & 42.1\\ 
DIMN~\cite{song2019generalizable} & 52.0 & 39.2 & 41.1 & 29.3 & 60.1 & 51.2 & 78.4 & 70.2 & 57.9 & 47.5\\ 
SNR~\cite{jin2020style} & 66.5 & 52.1 & 47.7 & 40.2 & 61.3 & 52.9 & 89.9 & 84.1 & 66.3 & 57.3\\ 
RaMoE~\cite{dai2021generalizable} & 67.3 & 57.7 & 54.2 & 46.8 & 64.6 & 56.6 & \textcolor[HTML]{FF0000}{\textbf{90.2}} & \textcolor[HTML]{FF0000}{\textbf{85.0}} & 69.1 & 61.5\\ 
\hline
\method{abbr}~(Ours) & \textcolor[HTML]{FF0000}{\textbf{72.2}} & \textcolor[HTML]{FF0000}{\textbf{62.7}} & \textcolor[HTML]{FF0000}{\textbf{59.4}} & \textcolor[HTML]{FF0000}{\textbf{48.4}} & \textcolor[HTML]{FF0000}{\textbf{70.1}} & \textcolor[HTML]{FF0000}{\textbf{61.2}} & 85.7 & 79.8 & \textcolor[HTML]{FF0000}{\textbf{71.9}} & \textcolor[HTML]{FF0000}{\textbf{63.0}}\\ 
\hline
\end{tabular}
  }
  \label{tab:sota_small}
\end{table}

\vspace{-1em}
\vparagraph{Comparison under large-scale benchmark setting.}
We further evaluated our model on four {\em large-scale} datasets
(M+D+C3+MS)
with the `leave-one-out' strategy,
namely taking three datasets used as source domains
for model training,
and one left out as an unseen target domain.
Under this setting,
The original train splits in the three source domains were used for training,
while the test split on the unseen target domain was used for testing,
same as in~\cite{zhao2021learning}.
The evaluation results in Table~\ref{tab:sota_large} show that
\method{abbr} outperforms
the SOTA competitors by a compelling margin,
Specially,
on the more challenging datasets
CUHK03 and MSMT17
with larger domain gaps to the other datasets, all
methods give relatively poorer generalization performances. In comparison, our \method{abbr} model
gains greater advantage over the other methods especially on
Rank-1 scores.
This suggests
\method{abbr}'s better scalability with greater potential
in real-world deployment to different unseen target domains.

\begin{table}[ht]
  \centering
  \caption{Comparisons with the SOTA
    generalized person ReID models
    on large-scale datasets.
  }
  \resizebox{\textwidth}{!}{
    \begin{tabular}{l|c|c|c|c|c|c|c|c|c|c}
\hline
 & \multicolumn{2}{c|}{Market-1501 } & \multicolumn{2}{c|}{DukeMTMC } & \multicolumn{2}{c|}{CUHK03} & \multicolumn{2}{c|}{MSMT17} & \multicolumn{2}{c}{Average}\\ 
\cline{2-11}
\multirow{-2}{*}{Method} & mAP & Rank-1 & mAP & Rank-1 & mAP & Rank-1 & mAP & Rank-1 & mAP & Rank-1\\ 
\hline
QAConv$_{50}$~\cite{liao2020interpretable} & 39.5 & 68.6 & 43.4 & 64.9 & 19.2 & 22.9 & 10.0 & 29.9 & 28.0 & 46.6\\ 
M3L~\cite{zhao2021learning} & 51.1 & 76.5 & 48.2 & 67.1 & 30.9 & 31.9 & 13.1 & 32.0 & 35.8 & 51.9\\ 
M3L(IBN)~\cite{zhao2021learning} & 52.5 & 78.3 & 48.8 & 67.2 & 31.4 & 31.6 & 15.4 & 37.1 & 37.0 & 53.5\\ 
\hline
\method{abbr}~(Ours) & \textcolor[HTML]{FF0000}{\textbf{58.3}} & \textcolor[HTML]{FF0000}{\textbf{81.4}} & \textcolor[HTML]{FF0000}{\textbf{49.8}} & \textcolor[HTML]{FF0000}{\textbf{70.0}} & \textcolor[HTML]{FF0000}{\textbf{34.1}} & \textcolor[HTML]{FF0000}{\textbf{35.5}} & \textcolor[HTML]{FF0000}{\textbf{17.7}} & \textcolor[HTML]{FF0000}{\textbf{43.1}} & \textcolor[HTML]{FF0000}{\textbf{40.0}} & \textcolor[HTML]{FF0000}{\textbf{57.5}}\\ 
\hline
\end{tabular}
  }
  \label{tab:sota_large}
  \vspace{-1em}
\end{table}

\subsection{Ablation Study}
\label{sec:ablation}

\vspace{-0.5em}
\vparagraph{Components analysis.}
We investigated the effects of different components
in \method{abbr} model design
to study individual contributions.
We trained a baseline model
with only identity loss $\mathcal{L}_\text{id}$,
and then incorporated it with either
LPM or GCM as well as both (\method{abbr}).
Table~\ref{tab:comp_abl}
shows that both the LPM and GCM
are beneficial individually,
and the benefits become clearer
when they are jointly adopted as in the \method{abbr} model.
From another perspective,
it also verifies that solely considering
the local or global regularization is biased,
and it is non-trivial that the \method{abbr} explores
both in a unified framework
to learn a more generic representation.

\begin{table}[ht]
  \centering
  \caption{Components analysis of LPM and GCM. \protect\method{abbr} incorporates both in a unified framework.}
  \resizebox{\textwidth}{!}{
    \begin{tabular}{l|c|c|c|c|c|c|c|c|c|c}
\hline
 & \multicolumn{2}{c|}{Market-1501 } & \multicolumn{2}{c|}{DukeMTMC } & \multicolumn{2}{c|}{CUHK03} & \multicolumn{2}{c|}{MSMT17} & \multicolumn{2}{c}{Average}\\ 
\cline{2-11}
\multirow{-2}{*}{Setting} & mAP & Rank-1 & mAP & Rank-1 & mAP & Rank-1 & mAP & Rank-1 & mAP & Rank-1\\ 
\hline
baseline & 54.1 & 78.5 & 49.0 & 68.1 & 31.1 & 31.9 & 14.9 & 38.1 & 37.3 & 54.1\\ 
+LPM & 57.9 & 80.4 & 49.4 & 69.4 & 32.7 & 33.2 & \textcolor[HTML]{FF0000}{\textbf{17.7}} & 42.8 & 39.4 & 56.5\\ 
+GCM & 55.0 & 79.5 & 49.0 & 68.5 & 32.6 & 33.6 & 16.1 & 39.4 & 38.2 & 55.2\\ 
\method{abbr} & \textcolor[HTML]{FF0000}{\textbf{58.3}} & \textcolor[HTML]{FF0000}{\textbf{81.4}} & \textcolor[HTML]{FF0000}{\textbf{49.8}} & \textcolor[HTML]{FF0000}{\textbf{70.0}} & \textcolor[HTML]{FF0000}{\textbf{34.1}} & \textcolor[HTML]{FF0000}{\textbf{35.5}} & \textcolor[HTML]{FF0000}{\textbf{17.7}} & \textcolor[HTML]{FF0000}{\textbf{43.1}} & \textcolor[HTML]{FF0000}{\textbf{40.0}} & \textcolor[HTML]{FF0000}{\textbf{57.5}}\\ 
\hline\hline
 & \multicolumn{2}{c|}{PRID} & \multicolumn{2}{c|}{GRID} & \multicolumn{2}{c|}{VIPeR} & \multicolumn{2}{c|}{iLIDs} & \multicolumn{2}{c}{Average}\\ 
\cline{2-11}
\multirow{-2}{*}{Setting} & mAP & Rank-1 & mAP & Rank-1 & mAP & Rank-1 & mAP & Rank-1 & mAP & Rank-1\\ 
\hline
baseline & 69.1 & 59.0 & 59.0 & 48.4 & 68.9 & 60.1 & 82.5 & 74.5 & 69.9 & 60.5\\ 
+LPM & 71.5 & 61.2 & 58.0 & \textcolor[HTML]{FF0000}{\textbf{48.5}} & 69.7 & 60.9 & 85.3 & 78.7 & 71.1 & 62.3\\ 
+GCM & 69.7 & 59.5 & 59.1 & \textcolor[HTML]{FF0000}{\textbf{48.5}} & 69.7 & 60.7 & 85.3 & 78.7 & 71.0 & 61.8\\ 
\method{abbr} & \textcolor[HTML]{FF0000}{\textbf{72.2}} & \textcolor[HTML]{FF0000}{\textbf{62.7}} & \textcolor[HTML]{FF0000}{\textbf{59.4}} & 48.4 & \textcolor[HTML]{FF0000}{\textbf{70.1}} & \textcolor[HTML]{FF0000}{\textbf{61.2}} & \textcolor[HTML]{FF0000}{\textbf{85.7}} & \textcolor[HTML]{FF0000}{\textbf{79.8}} & \textcolor[HTML]{FF0000}{\textbf{71.9}} & \textcolor[HTML]{FF0000}{\textbf{63.0}}\\ 
\cline{1-1}\cline{2-11}
\end{tabular}
  }
  \label{tab:comp_abl}
  \vspace{-0.5em}
\end{table}

\vspace{-0.5em}
\vparagraph{Discrimination and generalization trade-off.}
There is a trade-off
between being discriminative to the source domains,
and being generalized to the target domains~\cite{zhang2017discrimination}.
We quantitatively assessed the proposed \method{abbr} model
in this regard.
The results in Table~\ref{tab:abl_src_tgt}
indicate the baseline method
fails to generalize well to the target domains
but yielded compelling discrimination capacity
in the source domains,
which is likely due to overfitting.
As a comparison, our \method{abbr} gains
notable improvements in generalization ability
with only slight performance drops in the source domains.
This implies that \method{abbr} can
effectively balance the
generalization and discrimination
of feature representations,
so to be applied to any novel unseen domains.

\begin{table}[ht]
  \centering
  \caption{Local discrimination and global generalization trade-off.}
  \setlength{\tabcolsep}{3mm}
  \begin{tabular}{l|c|c|c|c|c|c|c|c}
\hline
 & \multicolumn{2}{c|}{Source Average} & \multicolumn{2}{c|}{Target: M} & \multicolumn{2}{c|}{Source Average} & \multicolumn{2}{c}{Target: D}\\ 
\cline{2-9}
\multirow{-2}{*}{Setting} & mAP & Rank-1 & mAP & Rank-1 & mAP & Rank-1 & mAP & Rank-1\\ 
\hline
baseline & \textcolor[HTML]{FF0000}{\textbf{58.6}} & \textcolor[HTML]{FF0000}{\textbf{73.8}} & 54.1 & 78.5 & \textcolor[HTML]{FF0000}{\textbf{63.0}} & \textcolor[HTML]{FF0000}{\textbf{77.4}} & 49.0 & 68.1\\ 
LPGC & 57.9 & 73.0 & \textcolor[HTML]{FF0000}{\textbf{58.3}} & \textcolor[HTML]{FF0000}{\textbf{81.4}} & 61.9 & 76.3 & \textcolor[HTML]{FF0000}{\textbf{49.8}} & \textcolor[HTML]{FF0000}{\textbf{70.0}}\\ 
\hline\hline
 & \multicolumn{2}{c|}{Source Average} & \multicolumn{2}{c|}{Target: C} & \multicolumn{2}{c|}{Source Average} & \multicolumn{2}{c}{Target: MS}\\ 
\cline{2-9}
\multirow{-2}{*}{Setting} & mAP & Rank-1 & mAP & Rank-1 & mAP & Rank-1 & mAP & Rank-1\\ 
\hline
baseline & \textcolor[HTML]{FF0000}{\textbf{64.2}} & \textcolor[HTML]{FF0000}{\textbf{82.3}} & 31.1 & 31.9 & \textcolor[HTML]{FF0000}{\textbf{69.9}} & \textcolor[HTML]{FF0000}{\textbf{79.3}} & 14.9 & 38.1\\ 
LPGC & 63.9 & 82.2 & \textcolor[HTML]{FF0000}{\textbf{34.1}} & \textcolor[HTML]{FF0000}{\textbf{35.5}} & 68.8 & 78.7 & \textcolor[HTML]{FF0000}{\textbf{17.7}} & \textcolor[HTML]{FF0000}{\textbf{43.1}}\\ 
\hline
\end{tabular}
  \label{tab:abl_src_tgt}
  \vspace{-0.5em}
\end{table}

\vparagraph{Effects of distribution perturbation on different layers.}

\begin{figure}[ht]
  \centering
  \vspace{-1.5em}
  \subfigure[Averaged performance under large-scale setting.]{
    \includegraphics[width=0.49\linewidth]{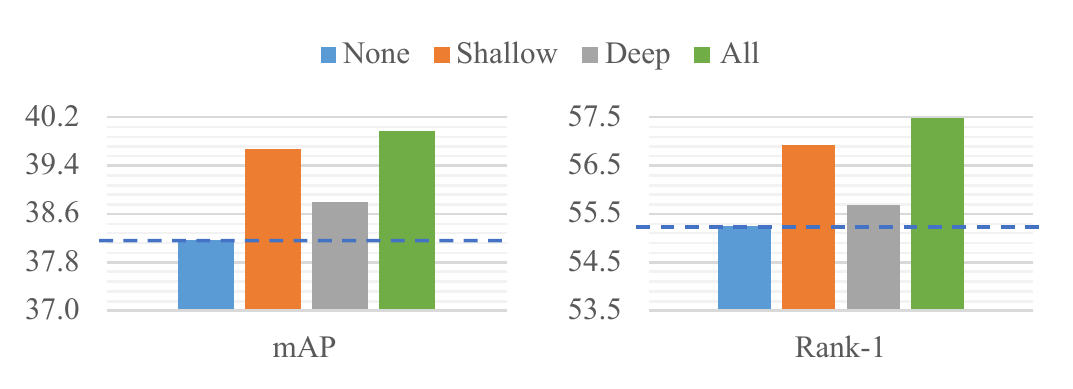}}
  \subfigure[Averaged performance under traditional setting.]{
    \includegraphics[width=0.49\linewidth]{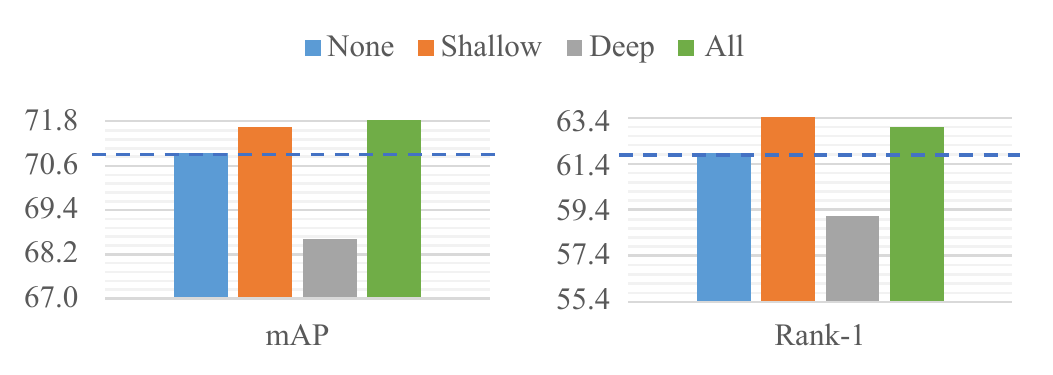}}
  \vspace{-1em}
  \caption{Effects of distribution perturbation on different layers.
  }
  \label{fig:local_loc}
  \vspace{-0.5em}
\end{figure}

We studied the effects of perturbing
the input distributions
of various layers in our backbone network,
including the `Shallow' layers
(the first convolution layer and a following residual block),
and `Deep' layers
(the last two residual blocks).
The results are shown in Figure~\ref{fig:local_loc}.
It is not a surprise that perturbing the shallow layers
consistently improves the performance
under both the traditional and large-scale settings,
as perturbations in earlier stages
helps enhance the invariance
of most layers to distribution shift.
However,
solely perturbing the deep layers
exhibit distinct behaviors under different benchmark settings.
This is because
the training data under the traditional setting
is relatively smaller
with restricted diversity
and
perturbations in later stages
tends to affect a limited part of the network
that is insufficient to
improve the model's generalization ability.
Based on these observations,
we propose to perturb all the layers
to improve the robustness of the \method{abbr} model
regardless of the dataset scale.

\vparagraph{Effects of the global calibration objective.}
\begin{wraptable}{r}{0.6\linewidth}
  \vspace{-1.8em}
  \centering
  \caption{Effects of the global calibration objective whose importance is decided by the weight $\lambda$ in Eq.~\ref{equ:loss_all}.
    Averaged performances are reported.
  }
  \vspace{0.5em}
  \resizebox{0.6\textwidth}{!}{
    \begin{tabular}{l|c|c|c|c}
\hline
 & \multicolumn{2}{c|}{Traditional setting} & \multicolumn{2}{c}{Large-scale setting}\\
\cline{2-5}
\multirow{-2}{*}{Setting} & mAP & Rank-1 & mAP & Rank-1\\
\hline
\method{abbr} (\textit{\textbf{default}, $\lambda=1$}) & \textcolor[HTML]{FF0000}{\textbf{71.9}} & \textcolor[HTML]{FF0000}{\textbf{63.0}} & \textcolor[HTML]{FF0000}{\textbf{40.0}} & \textcolor[HTML]{FF0000}{\textbf{57.5}}\\
\hline
\method{abbr} w/o GPM  & 71.0 & 61.9 & 37.5 & 54.3\\
\method{abbr} w/ $\lambda=0.1$  & 71.2 & 62.2 & 39.4 & 56.5\\
\method{abbr} w/ $\lambda=10$  & 70.8 & 62.1 & 39.2 & 56.5\\
\method{abbr} w/ $\lambda=100$  & 33.9 & 23.9 & 38.5 & 55.5\\
\hline
\end{tabular}
  }
  \vspace{-1em}
  \label{tab:gcm_weight}
\end{wraptable}

The importance of the global calibration objective
for avoiding the model from overfitting to source domains
is determined by the hyperparameter $\lambda$ in Eq.~\eqref{equ:loss_all}.
By linearly varying $\lambda$ from 0.1 to 100,
we observed from Table~\ref{tab:gcm_weight}
that moderately applying GCM (\eg 0.1 or 1)
is beneficial to \method{abbr}'s generalizability;
further increasing $\lambda$ to a larger value (\eg 10 or 100)
brings more harm than help.
This is because the learning process
is dominated by the calibration regularization
and the model can barely learn from the identity loss,
hence, the resulted feature is less discriminative.
We also observed that the traditional setting
is relatively more sensitive to $\lambda$,
as it holds much less training data
for learning a robust model,
and a similar phenomenon is shown in Figure~\ref{fig:local_loc}.
Given the above observations,
we set $\lambda=1$ in practice for our \method{abbr} model.

\section{Conclusions}
In this work,
we presented a novel \textit{\method{full}} (\method{abbr}) model
to learn generic yet discriminative representation
in multiple source domains
generalizable to arbitrary unseen target domains
for more accurate unseen domain person ReID.
\method{abbr} simultaneously conducts model regularization
on local per-domain feature-distribution and global
cross-domain feature-distribution in order to learn a better
domain-invariant feature space representation.
Benefited from the diverse features
synthesized by local perturbation,
\method{abbr} expands per-domain feature distribution to enable
more robust to domain shifts.
From the global calibration,
feature distributions of different domains
are represented and holistically referenced in a shared feature space
with their domain-specific data characteristics
(\ie mean and variance of feature distributions)
being ignored,
resulting in higher model generalizability.
Experiments on extensive ReID datasets
show the performance advantages
of the proposed \method{abbr} model
over a wide range of state-of-the-art competitors.
Extensive ablation studies further provided in-depth analysis
of the individual components designed in \method{abbr} model.

\clearpage
\bibliographystyle{plain}
\bibliography{PECA}
\clearpage
\end{document}